%% file: main.tex
\RequirePackage{etex}
\documentclass[letterpaper, 10pt, conference]{ieeeconf} 
\IEEEoverridecommandlockouts              
\usepackage{graphicx} 
\usepackage{cite}
\usepackage{xcolor}
\usepackage{amssymb}
\usepackage{booktabs}
\usepackage{float}
\usepackage{multirow}
\usepackage{array}
\usepackage{ulem}
\usepackage{marvosym}
\usepackage{amsmath}
\usepackage{arydshln}
\usepackage{fix-cm} 
\usepackage{needspace}

\newcolumntype{L}[1]{>{\raggedright\arraybackslash}p{#1}}
\newcolumntype{C}[1]{>{\centering\arraybackslash}p{#1}}
\newcolumntype{R}[1]{>{\raggedleft\arraybackslash}p{#1}}
\input{packages.tex}
\usepackage{hyperref}
\hypersetup{
    colorlinks=true,
    linkcolor=blue,
    filecolor=black,      
    urlcolor=purple,
    citecolor=purple
}

\makeatletter

\title{\LARGE \bf
ConViTac: Aligning Visual-Tactile Fusion with Contrastive Representations
}
\author{Zhiyuan Wu, Yongqiang Zhao, and Shan Luo
\thanks{Department of Engineering, King’s College London, Strand, London,
WC2R 2LS, United Kingdom, \{zhiyuan.1.wu, yongqiang.zhao, shan.luo\}@kcl.ac.uk.}
}

\begin{document}
	
\thispagestyle{empty}
\pagestyle{empty}

\maketitle

\begin{abstract}
Vision and touch are two fundamental sensory modalities for robots, offering complementary information that enhances perception and manipulation tasks. Previous research has attempted to jointly learn visual-tactile representations to extract more meaningful information. However, these approaches often rely on direct combination, such as feature addition and concatenation, for modality fusion, which tend to result in poor feature integration. In this paper, we propose ConViTac, a visual-tactile representation learning network designed to enhance the alignment of features during fusion using contrastive representations. Our key contribution is a Contrastive Embedding Conditioning (CEC) mechanism that leverages a contrastive encoder pretrained through self-supervised contrastive learning to project visual and tactile inputs into unified latent embeddings. These embeddings are used to couple visual-tactile feature fusion through cross-modal attention, aiming at aligning the unified representations and enhancing performance on downstream tasks. We conduct extensive experiments to demonstrate the superiority of ConViTac in real world over current state-of-the-art methods and the effectiveness of our proposed CEC mechanism, which improves accuracy by up to 12.0\% in material classification and grasping prediction tasks. More details are on our \href{https://georgewuzy.github.io/ConViTac-website/}{project website}.
\end{abstract}

\begin{figure*} [t!]
	\centering
    \vspace{0.6em}
	\includegraphics[width=0.79 \textwidth]{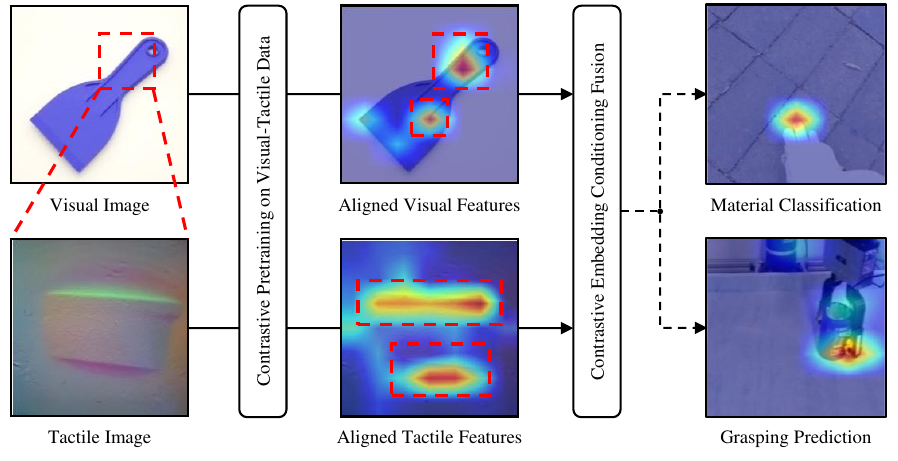}
    \vspace{-1.4em}
	\caption{
        An example of shovel in ObjectFolder Real \cite{gao2023objectfolderreal} for our ConViTac. Given a visual image and a corresponding tactile image, we obtain aligned visual-tactile features using contrastive pretraining on visual-tactile data, and leverage them for feature fusion through contrastive embedding conditioning (CEC) mechanism, which can be applied to substream tasks such as material classification and grasping prediction. 
    }\label{fig.intro}
    \vspace{-1.8em}
\end{figure*}

\section{Introduction} \label{introduction}
Robotic systems rely heavily on sensory modalities like vision and touch to perceive and interact with their environment effectively \cite{luo2017tactilesurvey3, chen2024zhuo1}. Vision provides a broad perception of the surroundings and is widely used in research. However, it often falls short in capturing dynamic and intricate states \cite{calandra2017calandra}. Tactile sensing, conversely, excels at detecting subtle features and capturing feedback that vision might overlook \cite{cao2020stam}. The integration of these modalities creates a robust perception system, with vision offering the environmental layout and configuration, while touch providing insights into dynamic interactions \cite{dave2024mvitac}. For example, in grasping tasks, vision helps in estimating the initial position and shape of objects, but it may be obstructed during the process, missing real-time changes like deformation and shifts \cite{luo2018vitac}. Here, tactile feedback adds valuable, dynamic interaction details that promote the estimation of pose and stability, enhancing the robot’s ability to perceive and understand its environment more comprehensively.

Recent advancements in vision-based tactile sensors have enabled the capture of more precise tactile information \cite{yuan2017gelsight, gomes2020geltip, cao2021touchroller}, and the structural similarity between vision-based tactile maps and visual data has led researchers to explore their integrated perception. Additionally, with the development of deep learning, the focus of visual-tactile fusion has shifted from data-level towards feature-level integration. Works such as \cite{calandra2017calandra, cui2020vtfsa} have explored supervised learning for visual-tactile feature fusion. However, these methods often combine features in a direct way, such as through concatenation or addition, which fail to associate tactile features with their corresponding parts in vision at the feature level. Alternatively, methods like \cite{yang2022touchandgo, kerr2022ssvtp, dave2024mvitac} have employed contrastive learning to obtain joint representations between the two modalities, but these are often trained in a self-supervised way as their objective is to learn the correspondence and similarity between visual-tactile data. When applied to downstream tasks, only a fully connected layer is trained with supervision, which weakens the effectiveness of the ground truth, as supervision cannot be effectively applied to the feature learning process.

The human brain demonstrates a remarkable ability to integrate visual and tactile information by visually pinpointing the area being touched and using tactile perceptions to enhance its understanding of that specific region within the visual field, aided by pre-learned semantic knowledge \cite{eimer2004neuroscience1}. In this paper, inspired by this idea in neuroscience, we introduce ConViTac, a novel visual-tactile representation learning method that improves multi-modal fusion through contrastive representations, due to their potentials to build connectivity between different modalities \cite{radford2021clip}. Central to our approach is the Contrastive Embedding Conditioning (CEC) mechanism, which integrates contrastive representations, originally obtained through self-supervised learning, into a \textbf{fully supervised} learning framework. CEC mechanism first train a contrastive encoder through SimCLR \cite{chen2020simclr} in a self-supervised way, and then leverages this pretrained contrastive encoder to project visual and tactile data into a learned joint space to get unified representations. These projected contrastive embeddings are then combined and used as a condition to align feature fusion via cross-modal attention, as illustrated in Fig. \ref{fig.intro}. Our experimental results demonstrate the superiority of ConViTac over existing state-of-the-art (SoTA) supervised and contrastive learning methods. We also perform ablation studies to quantitatively and qualitatively validate the effectiveness of the proposed CEC mechanism. 

The contributions can be summarized as follows:
\begin{itemize}
\item We propose ConViTac, a novel visual-tactile representation learning network that enhances feature alignment during fusion with contrastive representations. 
\item We propose a Contrastive Embedding Conditioning (CEC) mechanism to improve visual-tactile feature fusion through embedding projection and cross-modal attention. 
\item We conduct extensive experiments to validate the superiority of our ConViTac over SoTA baselines and demonstrate the effectiveness of our proposed CEC mechanism. 
\end{itemize}

\section{Related works} \label{relatedwork}
\subsection{Tactile Sensing}
Humans possess the ability to perceive physical properties such as hardness, roughness, and texture through tactile sensing \cite{chen2024zhuo2}, enabling effective feedback crucial for tasks like grasping and manipulation. Earlier tactile sensors mainly focused on fundamental, low-dimensional sensory outputs including force, pressure, vibration, and temperature \cite{luo2017tactilesurvey3}. Recently, however, significant advancements have been made in vision-based tactile sensors. GelSight \cite{yuan2017gelsight}, a notable example, achieves high-resolution tactile sensing by measuring surface deformations. Building upon it, advanced vision-based tactile sensors \cite{gomes2020geltip, cao2021touchroller} have been developed, offering extensive insights into object geometry, force, and shear. In parallel, research efforts have concentrated on innovative methods to utilize tactile information for robotic applications \cite{luo2017tactilesurvey3}. The majority of studies aim to incorporate tactile sensing to integrate the properties of objects into manipulation tasks, including material classification and grasping \cite{cao2020stam, luo2017tactilesurvey3}. For instance, numerous works utilize tactile feedback to enhance grasping stability \cite{schill2012grasping1}, which imposes high demands on the quality of tactile representations. In this paper, we introduce contrastive representations to improve the quality of tactile representations for a series of downstream tasks.

\subsection{Visual-Tactile Representation Learning}
There exists inherent connection between visual and tactile modalities: vision offers an overarching understanding of objects, while touch provides detailed insights \cite{dave2024mvitac}. The fusion of the two modalities can therefore yield a richer and more comprehensive informational content. With advancements in deep learning, recent studies on visual-tactile fusion have turned from data-level fusion to joint representation learning. Luo \textit{et al.} \cite{luo2018vitac} was the first to introduce feature fusion into visual-tactile representations for cloth texture recognition. Subsequent works \cite{cui2020vtfsa, chen2022vtt} have employed attention mechanisms to augment feature fusion between visual and tactile data. However, previous supervised visual-tactile learning networks mainly employ direct combination for fusion, such as feature addition and concatenation, which still leaves room for optimization.

\subsection{Multi-Modal Contrastive Learning}
The achievements of vision-language pretraining models \cite{radford2021clip} have underscored the potential of contrastive learning to bridge visual content and textual descriptions. Works like \cite{liu2021confusion1} have employed contrastive learning to improve feature fusion. In the field of visual-tactile learning, contrastive learning has emerged as a promising strategy, replacing traditional supervised learning. Techniques such as Information Noise Contrastive Estimation (InfoNCE) \cite{oord2018infonce} and Momentum Contrast (MoCo) \cite{he2020moco} have been utilized effectively to achieve joint representations between visual and tactile modalities, as demonstrated by studies including \cite{yang2022touchandgo, kerr2022ssvtp, dave2024mvitac}. Recently, UniTouch \cite{yang2024unitouch} leveraged the same contrastive learning to unify visual and tactile modalities, while additionally incorporating a language modality. Inspired by these prior works, this paper aims at employing contrastive representations to strengthen the coupling between visual and tactile representations within fully supervised learning.

\begin{figure*} [t!]
	\centering
    \vspace{0.6em}
	\includegraphics[width=0.99 \textwidth]{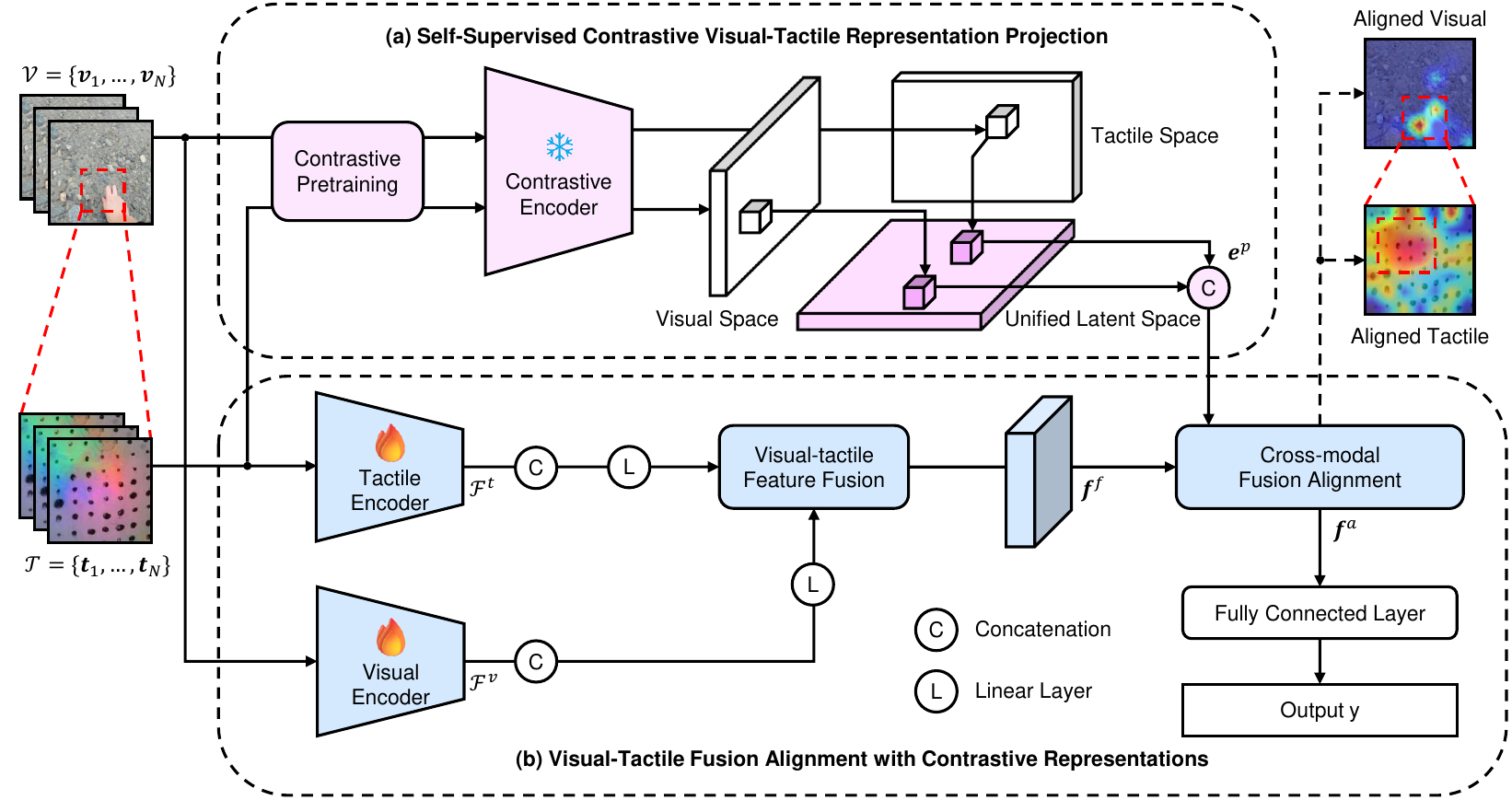}
    \vspace{-0.8em}
	\caption{
        An overview of our proposed ConViTac architecture. Our ConViTac processes visual and tactile sequences through a two-stage \textbf{C}ontrastive \textbf{E}mbedding \textbf{C}onditioning (CEC) mechanism: a) First, it performs self-supervised contrastive learning via SimCLR \cite{chen2020simclr} on all visual-tactile data to pretrain a contrastive encoder, which projects both visual and tactile inputs into a unified latent space. b) Subsequently, it employs the projected contrastive representations to align visual-tactile fusion in supervised learning, with the contrastive encoder frozen. The output $y$ represents the results from various substream tasks, including material classification and grasping prediction. Aligned features are visualized by GradCam \cite{selvaraju2017gradcam} for reference. 
    }\label{fig.pipeline}
    \vspace{-1.8em}
\end{figure*}

\section{Methodologies}\label{methodology}

\subsection{Architecture Overview}
Fig. \ref{fig.pipeline} depicts ConViTac, our proposed network which leverages contrastive representations to optimize feature fusion between vision and touch modalities. Since the visual and tactile sequences are captured using an RGB camera and a visual-based tactile sensor, the data can be formatted as images. We consider a visual sequence $\mathcal{V} = \{ \boldsymbol{v}_1, \dots, \boldsymbol{v}_{N} \}$ and a tactile sequence $\mathcal{T} = \{ \boldsymbol{t}_1, \dots, \boldsymbol{t}_{N} \}$, where $N$ represent the number of frames, and both sequences are of the same length, as visual and tactile data are collected in a synchronized manner with one-to-one correspondence in our setting. In the sequences, each element $\boldsymbol{v}_i$ or $\boldsymbol{t}_i$ is a tensor resized to the same dimension of $\mathbb{R}^{H \times W \times C}$ before being fed into networks, where $H, W$, and $C$ represent height, width, channel count, respectively. ConViTac consists of a) dual encoders to extract feature representations from $\mathcal{V}$ and $\mathcal{T}$, followed by a general fusion module $\oplus$ to integrate the extracted features; b) a contrastive encoder pretrained through self-supervised contrastive learning that projects both visual and tactile data into a shared latent space to obtain contrastive embeddings; and c) a cross-modal attention module to align fusion with the projected contrastive representations. Our network is fully supervised, utilizing cross-entropy loss for training.

\subsection{Visual-Tactile Encoders and Feature Fusion} \label{sec.fusion}
Our objective is to learn a comprehensive unified representation from the input sequences $\mathcal{V}$ and $\mathcal{T}$, which can be subsequently employed for downstream applications such as material classification and robotic manipulation. We utilize a Vision Transformer (ViT) \cite{dosovitskiy2020vit} as the backbone of our encoders. For $\mathcal{V}$, a sequence of visual features $\mathcal{F}^v = \{\boldsymbol{f}^v_1, \dots, \boldsymbol{f}^v_{N} \}$ is computed, and for $\mathcal{T}$, a sequence of tactile features $\mathcal{F}^t = \{ \boldsymbol{f}^t_1, \dots, \boldsymbol{f}^t_{N} \}$ is obtained. Both $\boldsymbol{f}^v_i$ and $\boldsymbol{f}^t_i$ belong to the same $\mathbb{R}^{P \times D}$, where $P$ indicates the number of patches and $D$ denotes the dimension of the feature embedding. 

While simple addition or concatenation serves as a baseline for feature fusion across modalities, our strategy focuses on exploring a more generalized paradigm to optimize the fusion process. By building on previous works that combine visual and tactile features \cite{cui2020vtfsa, luo2018vitac} and extending related work in computer vision \cite{wu2024s3mnet, wu2024sgroadseg} which harmonizes different modalities like RGB, depth, and normals, we concatenate $\mathcal{F}^v$ along dimension 0 to obtain feature maps with the dimension of $\mathbb{R}^{N \times P \times D}$, followed by a linear layer, and apply the same operation to $\mathcal{F}^t$. This process can be defined as:
\begin{equation} \label{eq.1}
    \boldsymbol{f}^f = \boldsymbol{L}_v [ \boldsymbol{C} ( \mathcal{F}^v ) ] \oplus \boldsymbol{L}_t [ \boldsymbol{C} ( \mathcal{F}^t ) ], 
\end{equation}
where $\boldsymbol{C}$ denotes the channel-wise concatenation, $\boldsymbol{L}_v$ and $\boldsymbol{L}_t$ represent linear projection operations that transform the concatenated visual and tactile feature maps, respectively. $\boldsymbol{f}^f \in \mathbb{R}^{2N \cdot P \cdot D}$ refers to the fused feature map, and $\oplus$ denotes a general feature fusion operation. 

\begin{table*}[t!]
        \fontsize{6.5}{10.0}\selectfont
	\centering
    \vspace{0.65em}
	\caption{
        \textbf{Material classification.} The metric is accuracy (\%) and the better results are in bold font. Chance refers to the baseline performance level that would be expected by a random classifier.
	}
    \vspace{-0.8em}
\label{tab.materialclassification}
	\begin{tabular}
            {C{2.0cm} L{1.7cm} C{0.5cm}C{0.5cm} C{1.4cm}C{1.4cm}C{1.5cm}C{1.4cm}C{1.4cm}C{1.5cm}}
            \toprule
            \multicolumn{1}{c}{\multirow{2}{*}{Category}} & \multicolumn{1}{c}{\multirow{2}{*}{Methods}} & \multicolumn{2}{c}{Modality} & \multicolumn{3}{c}{\textit{Touch and Go} } & \multicolumn{3}{c}{\textit{ObjectFolder Real} } \\
            &     & Vision & Touch & Category & Hard / Soft & Rough / Smooth & Category & Hard / Soft & Rough / Smooth \\
            \hline
            -& Chance &-&-& 18.6 & 66.1 & 56.3 & 13.8 & 50.6 & 49.0 \\
            \hline
            \multirow{3}{*}{\shortstack{Contrastive Learning}}& VT CMC \cite{yang2022touchandgo} & \checkmark & \checkmark & 68.6 & 87.1 & 82.4 & 29.5 & 59.8 & 65.6 \\
            & SSVTP \cite{kerr2022ssvtp} & \checkmark & \checkmark & 70.7 & 88.6 & 83.6 & 34.1 & 61.1 & 74.3 \\
            & MViTac \cite{dave2024mvitac} & \checkmark & \checkmark & 74.9 & 91.8 & 84.1 & 30.8 & 61.4 & 70.7 \\
            \hline
            \multirow{3}{*}{\shortstack{Supervised Learning}} 
            & STAM \cite{cao2020stam} &-& \checkmark & 52.6 & 88.9 & 75.1 & 40.4 & 67.0 & 68.6 \\
            & VTFSA \cite{cui2020vtfsa} & \checkmark & \checkmark & 66.8 & 92.5 & 82.2 & 47.9 & 72.2 & 74.1 \\
            \cdashline{2-10}
            & ConViTac (\textbf{Ours}) & \checkmark & \checkmark & \textbf{86.3} & \textbf{94.3} & \textbf{88.5} & \textbf{59.9} & \textbf{77.2} & \textbf{81.1} \\
    	\bottomrule
		\end{tabular}
    \vspace{-2.4em}
\end{table*}

\subsection{Aligning Feature Fusion with Contrastive Representations}
Contrastive learning has shown great potential in extracting shared latent representations across diverse modalities, such as vision and language \cite{radford2021clip}. This approach has been effectively utilized as a conditioning method to control tasks like image or 3D generation \cite{rombach2022stablediffusion, wu2025cdi3d}, which provides valuable insights for enhancing visual-tactile feature fusion. Drawing inspiration from these pioneering works, we propose a \textbf{C}ontrastive \textbf{E}mbedding \textbf{C}onditioning (CEC) mechanism that aligns visual-tactile feature fusion with contrastive representations to fully leverage the image-based nature of both modalities, given that our tactile data is acquired through vision-based tactile sensing. This mechanism is implemented through two steps: 1) self-supervised contrastive multi-modal representation projection and 2) cross-modal fusion alignment.

\subsubsection{Self-Supervised Contrastive Multi-Modal Representation Projection} \label{sec.contrastive_projection}

We begin by training a contrastive encoder $\mathcal{E}^c$ within all visual-tactile data using self-supervised contrastive learning to project visual and tactile data into latent embeddings within a shared feature space, which is achieved through SimCLR \cite{chen2020simclr}. During the self-supervised contrastive learning process, we employ $\mathcal{E}^c$ to obtain a projected latent embedding $\boldsymbol{e}^p \in \mathbb{R}^{2N \times P \times D}$  from any single image $\boldsymbol{v}$ in the visual sequence $\mathcal{V}$ and any single image $\boldsymbol{t}$ in the tactile sequence $\mathcal{T}$: 
\begin{equation}
    \boldsymbol{e}^p = \boldsymbol{C}[\mathcal{E}^c (\boldsymbol{v}), \mathcal{E}^c (\boldsymbol{t})]. 
\end{equation}

The loss function for the self-supervised learning $\mathcal{L}_c$ can be written as follows:  
\begin{equation}  
    \mathcal{L}_{c} = -\sum_{i=1}^{2B} \log \frac{\exp(S_{i,i+B})}{\sum_{j\neq i} \exp(S_{i,j})},  
\end{equation}  
where $B$ represents the batch size of all samples, and $S_{i,j}$ refers to the similarity matrix, calculated as:
\begin{equation}
    \boldsymbol{S}_{i, j} = \frac{\boldsymbol{e}^p_i \cdot \boldsymbol{e}^p_{j}}{\tau}, 
\end{equation}
where $\boldsymbol{e}^p_i$ represents the $i$-th projected feature in the batch, and $\tau$ represents the temperature parameter that scales the logits. We set the diagonal elements of $\boldsymbol{S}$ to $-\infty$ to prevent self-similarities. Notably, we utilize DINO \cite{zhang2022dino} as $\mathcal{E}^c$ given its superior performance based on ablation studies in Tab. \ref{tab.contrastive_encoder}. 

Consequently, we freeze this pretrained $\mathcal{E}^c$ during the following training, and utilize it to project both visual and tactile data into a shared feature space to get unified latent representations, as illustrated in Fig. \ref{fig.pipeline}. 

\subsubsection{Cross-modal Fusion Alignment}
We take the projected contrastive embeddings $\boldsymbol{e}^p$ as a condition to control the coupling in feature fusion through cross-modal attention. Given the fused feature $\boldsymbol{f}^f \in \mathbb{R}^{2N \cdot P \cdot D}$ from Section \ref{sec.fusion}, this alignment process can be formulated as follows:
\begin{equation}
    \boldsymbol{f}^a = \boldsymbol{C}[\mathcal{A}^{cm}_1(\boldsymbol{e}^p, \boldsymbol{f}^f), \dots, \mathcal{A}^{cm}_h(\boldsymbol{e}^p, \boldsymbol{f}^f)]\boldsymbol{w}_0, 
\end{equation}
where $\boldsymbol{f}^a$ refers to the aligned fused feature map, $h$ is the number of attention heads that empirically set to $8$, and $\boldsymbol{w}_0$ represents the weight matrix used for output. The cross-modal attention operation $\mathcal{A}^{cm}_i$ is defined as:
\begin{equation}
    \mathcal{A}_{cm}(\boldsymbol{e}, \boldsymbol{f}) = softmax(\frac{\boldsymbol{q}\boldsymbol{k}^T}{\sqrt{d}}) \cdot \boldsymbol{v}, 
\end{equation}
with 
\begin{equation}
    \boldsymbol{q} = \boldsymbol{w}_q \cdot \boldsymbol{e}, \quad \boldsymbol{k} = \boldsymbol{w}_k \cdot \boldsymbol{f}, \quad \boldsymbol{v} = \boldsymbol{w}_v \cdot \boldsymbol{f}, 
\end{equation}
where $\boldsymbol{w}$ are learnable projection matrices \cite{dosovitskiy2020vit}. 

\begin{table}[t!]
        \fontsize{6.5}{10.0}\selectfont
	\centering
    \vspace{0.65em}
	\caption{
        \textbf{Predicting success of grasping.} The metric is accuracy (\%) and the best results are in bold font. Chance refers to the baseline performance level that would be expected by a random classifier.
	}
    \vspace{-0.8em}
\label{tab.grasping}
	\begin{tabular}
            {c l C{0.4cm} C{0.4cm} c}
            \toprule
            \multicolumn{1}{c}{\multirow{2}{*}{Category}} & \multicolumn{1}{c}{\multirow{2}{*}{Methods}} & \multicolumn{2}{c}{Modality} & \multicolumn{1}{c}{\multirow{2}{*}{Accuracy (\%)}} \\
            & & Vision & Touch & \\
            \hline
            -& Chance &-&-& 50.8 \\
            \hline
            \multirow{2}{*}{\shortstack{Contrastive Learning}}
            & VT CMC \cite{yang2022touchandgo} & \checkmark & \checkmark & 56.3 \\
            & SSVTP \cite{kerr2022ssvtp} & \checkmark & \checkmark & 59.9 \\
            & MViTac \cite{dave2024mvitac} & \checkmark & \checkmark & 60.3 \\
            \hline
            \multirow{4}{*}{\shortstack{Supervised Learning}}
            & STAM \cite{cao2020stam} &-& \checkmark & 80.0 \\
            & Calandra \textit{et. al} \cite{calandra2017calandra} & \checkmark & \checkmark & 73.1 \\
            & VTFSA \cite{cui2020vtfsa} & \checkmark & \checkmark & 78.1 \\
            \cdashline{2-5}
            & ConViTac (\textbf{Ours}) & \checkmark & \checkmark & \textbf{84.3} \\
    	\bottomrule
		\end{tabular}
    \vspace{-2.4em}
\end{table}

\section{Experiments} \label{experimentresults}
This section details the experiments conducted to demonstrate the effectiveness of our proposed ConViTac model across four downstream tasks. We benchmark its performance against SoTA visual-tactile learning methods. Additionally, we perform ablation studies to validate the effectiveness of our CEC mechanism on different visual-tactile feature fusion strategies, both quantitatively and qualitatively.

\subsection{Datasets and Experimental Setups}
We evaluate our approach on the following tasks: 1) material property identification, specifically i) material classification, ii) discrimination between hard versus soft surfaces, and iii) distinction between smooth versus textured surfaces, and 2) robot grasping success prediction. The following datasets are utilized in our experiments:

\subsubsection{\textbf{Touch and Go dataset}}
The Touch and Go dataset \cite{yang2022touchandgo} is a recent, real-world visual-tactile collection obtained from interactions with various objects in diverse environments. It includes 13,900 tactile samples from around 4,000 distinct objects across 20 material categories. We follow the dataset splits from the original paper (70\% training, 30\% testing) to ensure consistency with previous studies. This dataset supports 20-class material categorization and binary classification tasks (hard/soft, rough/smooth).

\subsubsection{\textbf{ObjectFolder Real dataset}}
The ObjectFolder Real dataset \cite{gao2023objectfolderreal} provides a comprehensive multi-sensory depiction of real-world objects, including 3D meshes, video, impact sounds, and tactile data for 100 household items. These objects are categorized into 7 material classes, as well as binary classes (hard/soft, rough/smooth). We randomly split these objects into 70\% for training and 30\% for testing.

\subsubsection{\textbf{The Feeling of Success dataset}}
The Feeling of Success dataset \cite{calandra2017calandra} contains tactile sensor data from grippers along with RGB images, capturing samples at three stages: ‘before,’ ‘during,’ and ‘after’ an object is grasped. The goal is to predict the success of the grasp. Following \cite{dave2024mvitac}, we train models on 40 unique objects and evaluate on others, using only samples from the `during' stage.

We conduct our experiments on an NVIDIA RTX 3080Ti GPU, with a batch size of 16, using the Adam \cite{kingma2014adam} optimizer for model training. The initial learning rate was set at 0.1, with the models typically converging within 30 epochs for each task and dataset, and the number of patches $P$ was fixed at 16. For baseline implementations, we followed the original papers' specifications. Model performance was evaluated using accuracy (Acc).

\begin{figure*} [t!]
	\centering
    \vspace{0.6em}
	\includegraphics[width=0.99 \textwidth]{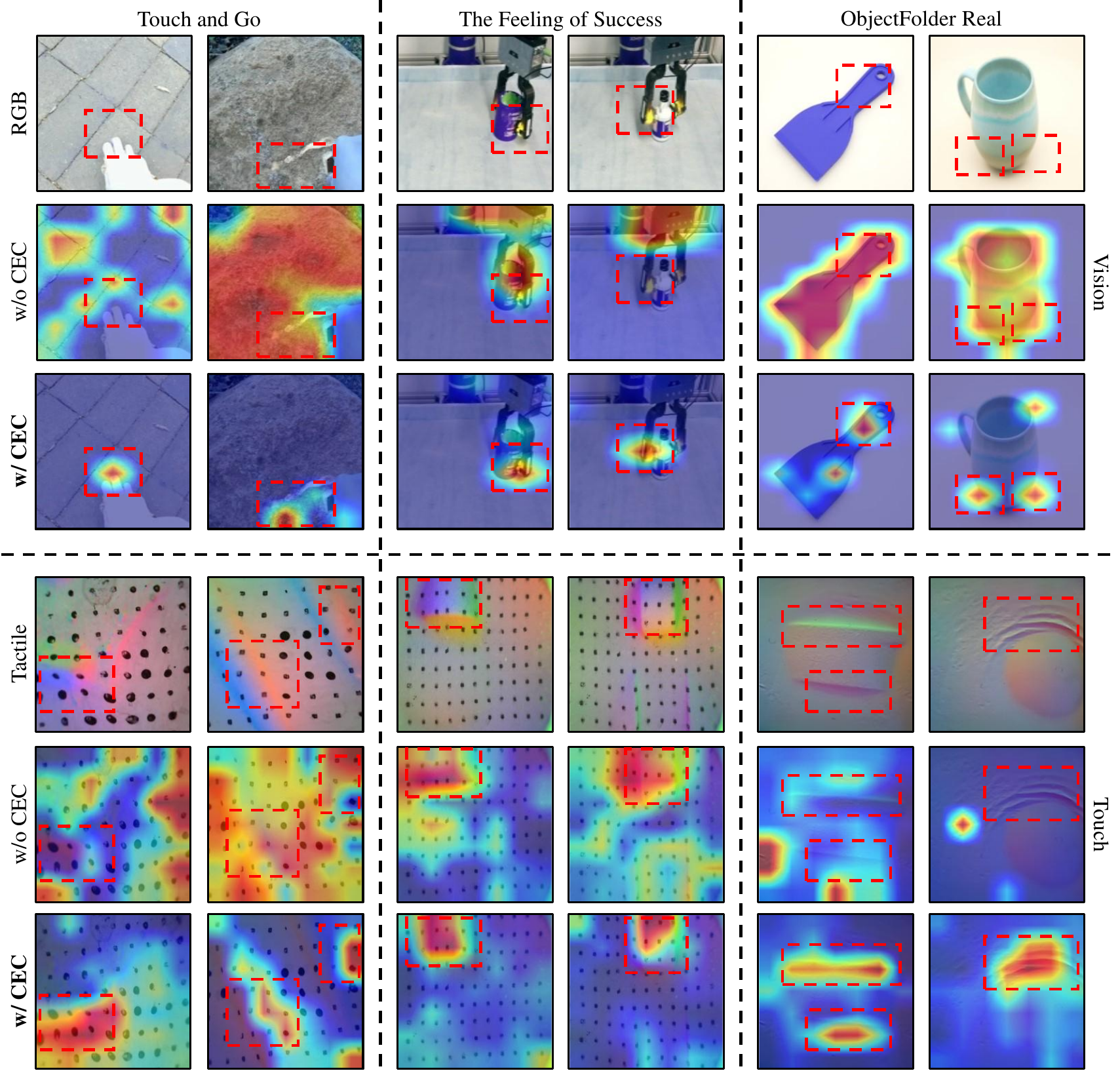}
    \vspace{-0.8em}
	\caption{
        Visualization for the effectiveness of our CEC mechanism using Grad-Cam \cite{selvaraju2017gradcam} with highlights in red boxes. With CEC mechanism, our ConViTac pays more attention to potential contacting areas for both visual and tactile modalities, promoting their connectivity and thereby enhancing the effectiveness of visual-tactile feature fusion. In visual data, ConViTac identifies touching areas in the Touch and Go dataset \cite{yang2022touchandgo}, grasp points in the Feeling of Success dataset \cite{calandra2017calandra}, and potential contact areas corresponding to tactile data within objects in the ObjectFolder dataset \cite{gao2023objectfolderreal}. For tactile data, ConViTac with CEC mechanism also focuses more precisely on contact areas.
    }\label{fig.viz}
    \vspace{-1.8em}
\end{figure*}

\subsection{Comparison with SoTA Methods}
\subsubsection{Material Classification} \label{exp.materialclassification}
We first evaluate ConViTac on material identification tasks using the Touch and Go \cite{yang2022touchandgo} and ObjectFolder Real \cite{gao2023objectfolderreal} datasets. Our comparisons include SoTA methods from both contrastive and supervised learning paradigms. For contrastive learning, we selected Visual-Tactile Contrastive Multiview Coding (VT CMC) \cite{yang2022touchandgo}, Self-Supervised Visuo-Tactile Pretraining (SSVTP) \cite{kerr2022ssvtp}, and Multimodal Visual-Tactile Representation Learning (MViTac) \cite{dave2024mvitac}. These models employ self-supervised contrastive loss functions during pretraining, after which encoders are frozen, and a linear classifier is trained with supervision to perform downstream tasks. Supervised methods included the Spatio-temporal Attention Model (STAM) \cite{cao2020stam} and the Visual-Tactile Fusion Self-Attention Model (VTFSA) \cite{cui2020vtfsa}. 

The quantitative experimental results presented in Table \ref{tab.materialclassification} indicate that contrastive learning methods surpass traditional supervised learning approaches on the Touch and Go dataset \cite{yang2022touchandgo}, probably due to the effective use of contrastive loss during the self-supervised pretraining phase. In contrast, traditional supervised methods generally employ duplex encoders for feature extraction and combine these features using conventional regression techniques, outperforming contrastive learning methods on the ObjectFolder Real dataset \cite{gao2023objectfolderreal}. Notably, our ConViTac achieves the best performance on both datasets. ConViTac outperforms baseline methods by 33.7\% to 11.4\% on the Touch and Go (category) dataset \cite{yang2022touchandgo} and 30.4\% to 12.0\% on the ObjectFolder Real (category) dataset \cite{gao2023objectfolderreal}. This superior performance is attributed to the projection of visual and tactile inputs into latent embeddings as a conditioned signal, which promotes visual-tactile feature fusion. Unlike traditional contrastive learning methods relying solely on loss functions for representation optimization, our methodology leverages a pretrained contrastive projection encoder, optimizing the modality representations at the feature level. Our approach applies contrastive supervision while maintaining cross-entropy loss for direct prediction error measurement, effectively combining the strengths of both supervised and contrastive learning paradigms to yield improved results.

\subsubsection{Predicting Grasping Success}
We evaluate our ConViTac model on predicting grasping success using the Feeling of Success dataset \cite{calandra2017calandra}. Following the same experimental setup outlined in Section \ref{exp.materialclassification}, we also include the supervised method from \cite{calandra2017calandra} for comparison. As shown in Table \ref{tab.grasping}, it can be observed that traditional supervised methods significantly outperform contrastive learning approaches for predicting the success of grasping unseen objects. Among the baseline methods, STAM \cite{cao2020stam} achieves superior accuracy, where only tactile modality is utilized as input. However, when integrated with contrastive conditioning, our ConViTac model achieves the best performance, demonstrating the effectiveness of leveraging contrastive representations to enhance visual-tactile integration. This highlights the potential of contrastive conditioning in harnessing subtle interactions between modalities, which are crucial for accurately predicting grasping outcomes in complex, unstructured environments.

\begin{table}[t!]
        \fontsize{6.5}{10.0}\selectfont
	\centering
    \vspace{0.65em}
	\caption{
        Comparison results of our proposed ConViTac with different contrastive encoder architectures. We test CNN \cite{he2016resnet}, ViT \cite{dosovitskiy2020vit}, and DINO \cite{zhang2022dino}. 
	}
    \vspace{-0.8em}
\label{tab.contrastive_encoder}
	\begin{tabular}
            { l c c c}
            \toprule
            \multicolumn{1}{c}{\multirow{2}{*}{Architectures}} & \multicolumn{3}{c}{Dataset} \\
            & \textit{Touch and Go} & \textit{The Feeling of Success} & \textit{ObjectFolder Real} \\
            \hline
            Baseline & 79.3 & 77.0 & 50.7 \\
            CNN \cite{he2016resnet} & 84.2 & 84.1 & 52.4 \\
            ViT \cite{dosovitskiy2020vit} & 84.3 & 83.9 & 56.5 \\
            DINO \cite{zhang2022dino} & \textbf{86.3} & \textbf{84.3} & \textbf{59.9} \\
            \bottomrule
		\end{tabular}
    \vspace{-2.4em}
\end{table}

\subsection{Ablation Study}

\subsubsection{Contrastive Encoder Architecture}
We first conduct ablation studies on the architecture of contrastive encoder $\mathcal{E}^c$. We compare CNN \cite{he2016resnet}, ViT \cite{dosovitskiy2020vit}, and DINO \cite{zhang2022dino}. As demonstrated in Tab. \ref{tab.contrastive_encoder}, DINO outperforms traditional ViT and CNN architectures due to its superior visual representation capability, as it automatically attends to semantically relevant regions and learns more robust feature representations through self-supervised learning \cite{zhang2022dino}. 

It is also noteworthy that the implementation of the Contrastive Encoder Component (CEC) mechanism resulted in an increase of 91.79 MiB in our network's parameters, raising the total from 168.07 MiB to 259.86 MiB, which represents a \textbf{35.4\%} increase. Additionally, the processing speed experienced a decrease from 38.17 FPS to 31.85 FPS, reflecting a \textbf{16.6\%} reduction. Despite these changes, both the parameter increase and the processing speed reduction remain within acceptable limits, ensuring the network retains its real-time capability for practical applications.

\begin{table}[t!]
        \fontsize{6.5}{10.0}\selectfont
	\centering
	\caption{
        Ablation study on contrastive conditioning modalities. The metric is accuracy (\%) and the best results are in bold font. 
	}
    \vspace{-0.8em}
\label{tab.condition}
	\begin{tabular}
            {l c c c}
            \toprule
            \multicolumn{1}{c}{Conditioning} 
            & \multicolumn{3}{c}{Dataset} \\
            \multicolumn{1}{c}{Modality}
            & \textit{Touch and Go} & \textit{The Feeling of Success} & \textit{ObjectFolder Real} \\
            \hline
            No Condition & 79.3 & 77.0 & 50.7 \\
            Vision & 84.4 & 82.8 & 56.1 \\
            Touch & 85.0 & 82.2 & 55.0 \\
            Vision + Touch & \textbf{86.3} & \textbf{84.3} & \textbf{59.9} \\
    	\bottomrule
		\end{tabular}
    \vspace{-2.4em}
\end{table}

\subsubsection{Contrastive Conditioning Modality}
We further analyze the effectiveness of contrastive conditioning modalities for latent embedding projection as described in Sec. \ref{sec.contrastive_projection}. To ensure generality, we employ concatenation for the feature fusion operation $\oplus$ in Eq. \ref{eq.1}. The results, presented in Table \ref{tab.condition}, demonstrate that our baseline network, absent of contrastive conditioning, underperforms, yielding accuracy inferior to that of STAM \cite{cao2020stam} on the Feeling of Success dataset \cite{calandra2017calandra}. In contrast, the introduction of contrastive conditioning significantly elevates performance. Employing either visual or tactile contrastive projection as conditioning yields improvements ranging from 4.3\% to 5.8\%. Further enhancement, from 1.3\% to 4.9\%, is observed when using concatenated visual and tactile contrastive embeddings.

\begin{figure} [t!]
	\centering
    \vspace{0.6em}
	\includegraphics[width=0.48 \textwidth]{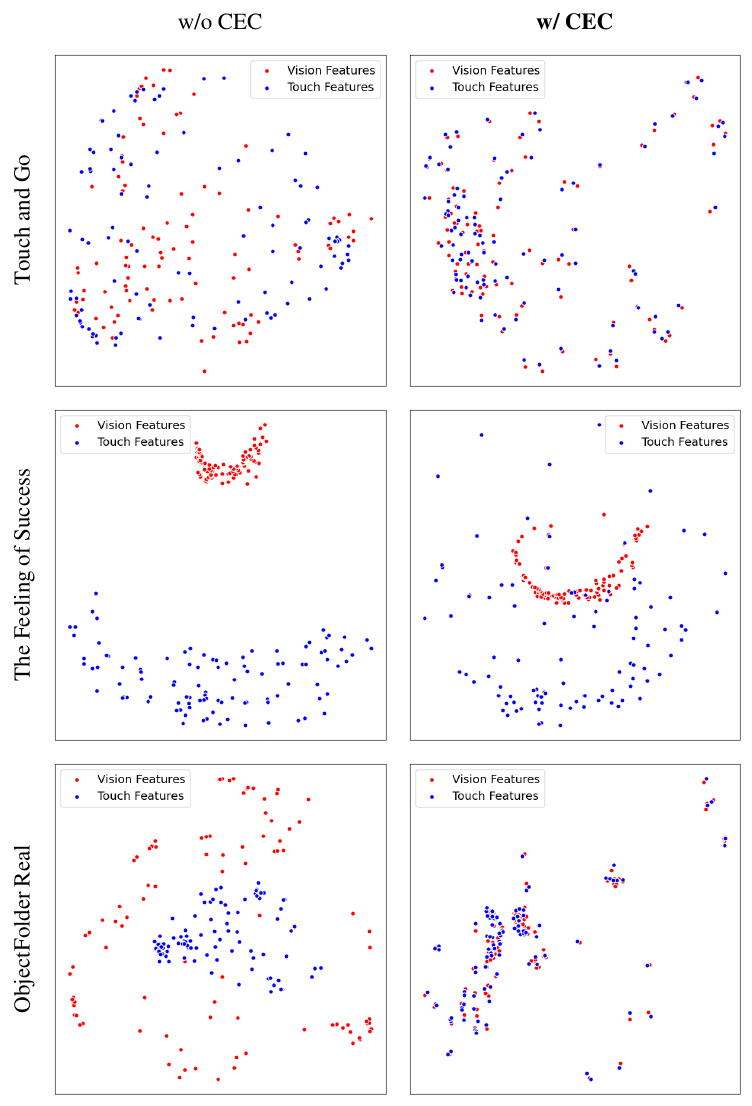}
    \vspace{-1.8em}
	\caption{
        Visualization for visual and tactile feature distribution with and without our CEC mechanism. We employ PCA \cite{yang2004pca} to reduce the dimensionality of visual and tactile features to project them into a two-dimensional coordinate system.
    }\label{fig.pca}
    \vspace{-0.8em}
\end{figure}

\begin{table}[t!]
        \fontsize{6.5}{10.0}\selectfont
	\centering
    \vspace{0.65em}
	\caption{
        Comparison results of our proposed ConViTac over visual-tactile fusion module. The metric is accuracy (\%) and the better results are in bold font. 
	}
    \vspace{-0.8em}
\label{tab.fusion}
	\begin{tabular}
            {l c c c}
            \toprule
            \multicolumn{1}{c}{\multirow{2}{*}{Fusion Modules}} & \multicolumn{3}{c}{Dataset} \\
            & \textit{Touch and Go} & \textit{The Feeling of Success} & \textit{ObjectFolder Real} \\
            \hline
            Add & 77.5 & 77.4 & 48.9 \\
            Add-\textbf{Con} & \textbf{80.8} & \textbf{82.5} & \textbf{56.3} \\
            \hline
            Concat & 79.3 & 77.0 & 50.7 \\
            Concat-\textbf{Con} & \textbf{86.3} & \textbf{84.3} & \textbf{59.9} \\
            \hline
            SWS \cite{wang2020sws1} & 78.2 & 78.4 & 48.3 \\
            SWS-\textbf{Con} & \textbf{82.7} & \textbf{81.9} & \textbf{53.7} \\
    	\bottomrule
		\end{tabular}
    \vspace{-2.4em}
\end{table}

\subsubsection{Alignment on Feature Fusion}
We also conduct ablation studies to validate the effectiveness of our CEC mechanism in aligning visual-tactile feature fusion strategies. We adopt three fusion operations for our ViT backbone: Addition, Concatenation, and the Softmax Weighted Sum (SWS), a commonly used approach in ViT-based fusion frameworks \cite{wang2020sws1}. Table \ref{tab.fusion} demonstrates the enhancement provided by CEC mechanism, highlighting improved accuracy across all fusion strategies. CEC mechanism substantially improves fusion strategies with accuracy increases of 3.3\% to 9.2\%. 

\subsubsection{Visualization for Contrastive Conditioning}
We visualize the efficacy of our CEC mechanism qualitatively utilizing Grad-Cam \cite{selvaraju2017gradcam}, and the heat-maps are illustrated in Fig. \ref{fig.viz}, demonstrating that with CEC mechanism, our ConViTac module focuses more accurately on potential contact areas. To be specific, without CEC mechanism, the networks broadly perceive entire objects or regions (the 2nd and 4th row). Specifically, for visual data, the network detects the robotic arm in the Feeling of Success dataset \cite{calandra2017calandra} and entire objects in the ObjectFolder Real dataset \cite{gao2023objectfolderreal}. For tactile data, it imprecisely identifies contact areas in the Feeling of Success dataset \cite{calandra2017calandra} and fails to identify informative areas in both the Touch and Go \cite{yang2022touchandgo} and ObjectFolder Real datasets \cite{gao2023objectfolderreal}. On the other hand, with the utilization of CEC mechanism, ConViTac effectively focuses on potential contacting areas (the 3rd and 6th row). In visual data, ConViTac identifies touching areas in the Touch and Go dataset \cite{yang2022touchandgo}, grasp points in the Feeling of Success dataset \cite{calandra2017calandra}, and potential contact areas corresponding to tactile data within objects in the ObjectFolder dataset \cite{gao2023objectfolderreal}. For tactile data, ConViTac with CEC mechanism also focuses more precisely on contact areas than the original network. 

Following \cite{cao2024pcaguan}, we also visualize the effectiveness of our CEC mechanism quantitatively by employing Principal Component Analysis (PCA) \cite{yang2004pca} to visualize the distribution of visual and tactile features before feature fusion. As illustrated in Fig. \ref{fig.pca}, it can be observed that with CEC mechanism, the network presents a more consistent distribution, indicating a higher degree of feature alignment and integration. This is more obvious in the Feeling of Success \cite{calandra2017calandra} and ObjectFolder Real datasets \cite{gao2023objectfolderreal}, which is corresponding to Fig. \ref{fig.viz}. This more similar and closer distribution of features suggests that our CEC mechanism effectively enhances the correlation between visual and tactile modalities with contrastive representations, leading to improved feature fusion. 

With visual and tactile embeddings as conditions, we promote visual-tactile feature fusion through cross-modal attention instead of simply combining them. This approach reorganizes the distribution of visual and tactile features, as well as fine-tuning the focus on interactive regions rich with informative features relevant to sub-tasks in both domains, which enhances performance in feature fusion and the execution of downstream tasks.

\section{Conclusion}\label{conclusion}
In this paper, we introduce ConViTac, a novel visual-tactile representation learning framework designed to align visual-tactile fusion with contrastive representations. To be specific, we present the Contrastive Embedding Conditioning (CEC) mechanism that leverages a contrastive encoder pretrained through self-supervised contrastive learning to project visual and tactile inputs into unified latent embeddings. These embeddings are used to couple visual-tactile feature fusion through cross-modal attention, aiming at aligning the unified representations and enhancing performance on downstream tasks. We conducted extensive experiments on material identification and grasping prediction datasets, demonstrating the superiority of ConViTac over SoTA baselines. Additionally, ablation studies confirmed the effectiveness of our CEC mechanism both qualitatively and quantitatively. In future work, we aim to extend the application of contrastive representations to more complex robotic learning tasks, such as peg insertion and lock opening.

\normalem
\bibliographystyle{IEEEtran}
\bibliography{egbib}

\end{document}

%% file: packages.tex
\usepackage{scalerel}
\usepackage{tikz}
\usetikzlibrary{svg.path}
\definecolor{orcidlogocol}{HTML}{A6CE39}
\tikzset{
  orcidlogo/.pic={
    \fill[orcidlogocol] svg{M256,128c0,70.7-57.3,128-128,128C57.3,256,0,198.7,0,128C0,57.3,57.3,0,128,0C198.7,0,256,57.3,256,128z};
    \fill[white] svg{M86.3,186.2H70.9V79.1h15.4v48.4V186.2z}
                 svg{M108.9,79.1h41.6c39.6,0,57,28.3,57,53.6c0,27.5-21.5,53.6-56.8,53.6h-41.8V79.1z M124.3,172.4h24.5c34.9,0,42.9-26.5,42.9-39.7c0-21.5-13.7-39.7-43.7-39.7h-23.7V172.4z}
                 svg{M88.7,56.8c0,5.5-4.5,10.1-10.1,10.1c-5.6,0-10.1-4.6-10.1-10.1c0-5.6,4.5-10.1,10.1-10.1C84.2,46.7,88.7,51.3,88.7,56.8z};
  }
}
\newcommand\orcidicon[1]{\href{https://orcid.org/#1}{\mbox{\scalerel*{
\begin{tikzpicture}[yscale=-1,transform shape]
\pic{orcidlogo};
\end{tikzpicture}
}{|}}}}
\usepackage{hyperref}
\hypersetup{
    colorlinks=true,
    linkcolor=blue,
    filecolor=black,      
    urlcolor=black,
    citecolor=purple
}
\usepackage{tikz-network}

%% file: main.bbl
\begin{thebibliography}{10}
\providecommand{\url}[1]{#1}
\csname url@samestyle\endcsname
\providecommand{\newblock}{\relax}
\providecommand{\bibinfo}[2]{#2}
\providecommand{\BIBentrySTDinterwordspacing}{\spaceskip=0pt\relax}
\providecommand{\BIBentryALTinterwordstretchfactor}{4}
\providecommand{\BIBentryALTinterwordspacing}{\spaceskip=\fontdimen2\font plus
\BIBentryALTinterwordstretchfactor\fontdimen3\font minus \fontdimen4\font\relax}
\providecommand{\BIBforeignlanguage}[2]{{%
\expandafter\ifx\csname l@#1\endcsname\relax
\typeout{** WARNING: IEEEtran.bst: No hyphenation pattern has been}%
\typeout{** loaded for the language `#1'. Using the pattern for}%
\typeout{** the default language instead.}%
\else
\language=\csname l@#1\endcsname
\fi
#2}}
\providecommand{\BIBdecl}{\relax}
\BIBdecl

\bibitem{gao2023objectfolderreal}
R.~Gao, Y.~Dou, H.~Li, T.~Agarwal, J.~Bohg, Y.~Li, L.~Fei-Fei, and J.~Wu, ``The objectfolder benchmark: Multisensory learning with neural and real objects,'' in \emph{Proceedings of the IEEE/CVF Conference on Computer Vision and Pattern Recognition (CVPR)}, 2023, pp. 17\,276--17\,286.

\bibitem{luo2017tactilesurvey3}
S.~Luo, J.~Bimbo, R.~Dahiya, and H.~Liu, ``Robotic tactile perception of object properties: A review,'' \emph{Mechatronics}, vol.~48, pp. 54--67, 2017.

\bibitem{chen2024zhuo1}
Z.~Chen, N.~Ou, J.~Jiang, and S.~Luo, ``Deep domain adaptation regression for force calibration of optical tactile sensors,'' in \emph{2024 IEEE/RSJ International Conference on Intelligent Robots and Systems (IROS)}.\hskip 1em plus 0.5em minus 0.4em\relax IEEE, 2024, pp. 13\,561--13\,568.

\bibitem{calandra2017calandra}
R.~Calandra, A.~Owens, M.~Upadhyaya, W.~Yuan, J.~Lin, E.~H. Adelson, and S.~Levine, ``The feeling of success: Does touch sensing help predict grasp outcomes?'' in \emph{Conference on Robot Learning (CoRL)}.\hskip 1em plus 0.5em minus 0.4em\relax PMLR, 2017, pp. 314--323.

\bibitem{cao2020stam}
G.~Cao, Y.~Zhou, D.~Bollegala, and S.~Luo, ``Spatio-temporal attention model for tactile texture recognition,'' in \emph{2020 IEEE/RSJ International Conference on Intelligent Robots and Systems (IROS)}.\hskip 1em plus 0.5em minus 0.4em\relax IEEE, 2020, pp. 9896--9902.

\bibitem{dave2024mvitac}
V.~Dave, F.~Lygerakis, and E.~Rueckert, ``Multimodal visual-tactile representation learning through self-supervised contrastive pre-training,'' \emph{arXiv preprint arXiv:2401.12024}, 2024.

\bibitem{luo2018vitac}
S.~Luo, W.~Yuan, E.~Adelson, A.~G. Cohn, and R.~Fuentes, ``Vitac: Feature sharing between vision and tactile sensing for cloth texture recognition,'' in \emph{2018 IEEE International Conference on Robotics and Automation (ICRA)}.\hskip 1em plus 0.5em minus 0.4em\relax IEEE, 2018, pp. 2722--2727.

\bibitem{yuan2017gelsight}
W.~Yuan, S.~Dong, and E.~H. Adelson, ``Gelsight: High-resolution robot tactile sensors for estimating geometry and force,'' \emph{Sensors}, vol.~17, no.~12, p. 2762, 2017.

\bibitem{gomes2020geltip}
D.~F. Gomes, Z.~Lin, and S.~Luo, ``Geltip: A finger-shaped optical tactile sensor for robotic manipulation,'' in \emph{2020 IEEE/RSJ International Conference on Intelligent Robots and Systems (IROS)}.\hskip 1em plus 0.5em minus 0.4em\relax IEEE, 2020, pp. 9903--9909.

\bibitem{cao2021touchroller}
G.~Cao, J.~Jiang, C.~Lu, D.~F. Gomes, and S.~Luo, ``Touchroller: A rolling optical tactile sensor for rapid assessment of large surfaces,'' \emph{arXiv preprint arXiv:2103.00595}, 2021.

\bibitem{cui2020vtfsa}
S.~Cui, R.~Wang, J.~Wei, J.~Hu, and S.~Wang, ``Self-attention based visual-tactile fusion learning for predicting grasp outcomes,'' \emph{IEEE Robotics and Automation Letters}, vol.~5, no.~4, pp. 5827--5834, 2020.

\bibitem{yang2022touchandgo}
F.~Yang, C.~Ma, J.~Zhang, J.~Zhu, W.~Yuan, and A.~Owens, ``Touch and go: Learning from human-collected vision and touch,'' \emph{Advances in Neural Information Processing Systems (NeurIPS)}, vol.~35, pp. 8081--8103, 2022.

\bibitem{kerr2022ssvtp}
J.~Kerr, H.~Huang, A.~Wilcox, R.~Hoque, J.~Ichnowski, R.~Calandra, and K.~Goldberg, ``Self-supervised visuo-tactile pretraining to locate and follow garment features,'' \emph{arXiv preprint arXiv:2209.13042}, 2022.

\bibitem{eimer2004neuroscience1}
M.~Eimer, ``Multisensory integration: How visual experience shapes spatial perception,'' \emph{Current biology}, vol.~14, no.~3, pp. R115--R117, 2004.

\bibitem{radford2021clip}
A.~Radford, J.~W. Kim, C.~Hallacy, A.~Ramesh, G.~Goh, S.~Agarwal, G.~Sastry, A.~Askell, P.~Mishkin, J.~Clark \emph{et~al.}, ``Learning transferable visual models from natural language supervision,'' in \emph{International conference on machine learning (ICML)}.\hskip 1em plus 0.5em minus 0.4em\relax PMLR, 2021, pp. 8748--8763.

\bibitem{chen2020simclr}
T.~Chen, S.~Kornblith, M.~Norouzi, and G.~Hinton, ``A simple framework for contrastive learning of visual representations,'' in \emph{International conference on machine learning}.\hskip 1em plus 0.5em minus 0.4em\relax PmLR, 2020, pp. 1597--1607.

\bibitem{chen2024zhuo2}
Z.~Chen, N.~Ou, X.~Zhang, and S.~Luo, ``Transforce: Transferable force prediction for vision-based tactile sensors with sequential image translation,'' \emph{arXiv preprint arXiv:2409.09870}, 2024.

\bibitem{schill2012grasping1}
J.~Schill, J.~Laaksonen, M.~Przybylski, V.~Kyrki, T.~Asfour, and R.~Dillmann, ``Learning continuous grasp stability for a humanoid robot hand based on tactile sensing,'' in \emph{2012 4th IEEE RAS \& EMBS International Conference on Biomedical Robotics and Biomechatronics (BioRob)}.\hskip 1em plus 0.5em minus 0.4em\relax IEEE, 2012, pp. 1901--1906.

\bibitem{chen2022vtt}
Y.~Chen, M.~Van~der Merwe, A.~Sipos, and N.~Fazeli, ``Visuo-tactile transformers for manipulation,'' in \emph{6th Annual Conference on Robot Learning}, 2022.

\bibitem{liu2021confusion1}
Y.~Liu, Q.~Fan, S.~Zhang, H.~Dong, T.~Funkhouser, and L.~Yi, ``Contrastive multimodal fusion with tupleinfonce,'' in \emph{Proceedings of the IEEE/CVF International Conference on Computer Vision}, 2021, pp. 754--763.

\bibitem{oord2018infonce}
A.~v.~d. Oord, Y.~Li, and O.~Vinyals, ``Representation learning with contrastive predictive coding,'' \emph{arXiv preprint arXiv:1807.03748}, 2018.

\bibitem{he2020moco}
K.~He, H.~Fan, Y.~Wu, S.~Xie, and R.~Girshick, ``Momentum contrast for unsupervised visual representation learning,'' in \emph{Proceedings of the IEEE/CVF conference on computer vision and pattern recognition}, 2020, pp. 9729--9738.

\bibitem{yang2024unitouch}
F.~Yang, C.~Feng, Z.~Chen, H.~Park, D.~Wang, Y.~Dou, Z.~Zeng, X.~Chen, R.~Gangopadhyay, A.~Owens \emph{et~al.}, ``Binding touch to everything: Learning unified multimodal tactile representations,'' in \emph{Proceedings of the IEEE/CVF Conference on Computer Vision and Pattern Recognition}, 2024, pp. 26\,340--26\,353.

\bibitem{selvaraju2017gradcam}
R.~R. Selvaraju, M.~Cogswell, A.~Das, R.~Vedantam, D.~Parikh, and D.~Batra, ``Grad-cam: Visual explanations from deep networks via gradient-based localization,'' in \emph{Proceedings of the IEEE international conference on computer vision}, 2017, pp. 618--626.

\bibitem{dosovitskiy2020vit}
A.~Dosovitskiy, L.~Beyer, A.~Kolesnikov, D.~Weissenborn, X.~Zhai, T.~Unterthiner, M.~Dehghani, M.~Minderer, G.~Heigold, S.~Gelly \emph{et~al.}, ``An image is worth 16x16 words: Transformers for image recognition at scale,'' in \emph{International Conference on Learning Representations (ICLR)}, 2020.

\bibitem{wu2024s3mnet}
Z.~Wu, Y.~Feng, C.-W. Liu, F.~Yu, Q.~Chen, and R.~Fan, ``S$^3$m-net: Joint learning of semantic segmentation and stereo matching for autonomous driving,'' \emph{IEEE Transactions on Intelligent Vehicles}, 2024.

\bibitem{wu2024sgroadseg}
Z.~Wu, J.~Li, Y.~Feng, C.~Liu, W.~Ye, Q.~Chen, and R.~Fan, ``Sg-roadseg: End-to-end collision-free space detection sharing encoder representations jointly learned via unsupervised deep stereo,'' in \emph{2024 International Conference on Robotics and Automation (ICRA)}.\hskip 1em plus 0.5em minus 0.4em\relax IEEE, 2024, p. in press.

\bibitem{rombach2022stablediffusion}
R.~Rombach, A.~Blattmann, D.~Lorenz, P.~Esser, and B.~Ommer, ``High-resolution image synthesis with latent diffusion models,'' in \emph{Proceedings of the IEEE/CVF conference on computer vision and pattern recognition}, 2022, pp. 10\,684--10\,695.

\bibitem{wu2025cdi3d}
Z.~Wu, X.~Song, S.~Wang, W.~Liu, J.~Yang, Z.~Cheng, S.~Chen, T.~Shang, W.~Sun, S.~Luo \emph{et~al.}, ``Cdi3d: Cross-guided dense-view interpolation for 3d reconstruction,'' \emph{arXiv preprint arXiv:2503.08005}, 2025.

\bibitem{zhang2022dino}
H.~Zhang, F.~Li, S.~Liu, L.~Zhang, H.~Su, J.~Zhu, L.~M. Ni, and H.-Y. Shum, ``Dino: Detr with improved denoising anchor boxes for end-to-end object detection,'' \emph{arXiv preprint arXiv:2203.03605}, 2022.

\bibitem{kingma2014adam}
D.~P. Kingma and J.~Ba, ``Adam: A method for stochastic optimization,'' \emph{arXiv preprint arXiv:1412.6980}, 2014.

\bibitem{he2016resnet}
K.~He, X.~Zhang, S.~Ren, and J.~Sun, ``Deep residual learning for image recognition,'' in \emph{Proceedings of the IEEE conference on computer vision and pattern recognition}, 2016, pp. 770--778.

\bibitem{yang2004pca}
J.~Yang, D.~Zhang, A.~F. Frangi, and J.-y. Yang, ``Two-dimensional pca: a new approach to appearance-based face representation and recognition,'' \emph{IEEE transactions on pattern analysis and machine intelligence}, vol.~26, no.~1, pp. 131--137, 2004.

\bibitem{wang2020sws1}
Y.~Wang, F.~Sun, M.~Lu, and A.~Yao, ``Learning deep multimodal feature representation with asymmetric multi-layer fusion,'' in \emph{Proceedings of the 28th ACM International Conference on Multimedia (ACM MM)}, 2020, pp. 3902--3910.

\bibitem{cao2024pcaguan}
G.~Cao, J.~Jiang, D.~Bollegala, M.~Li, and S.~Luo, ``Multimodal zero-shot learning for tactile texture recognition,'' \emph{Robotics and Autonomous Systems}, vol. 176, p. 104688, 2024.

\end{thebibliography}
